\pdfoutput=1

\documentclass[11pt]{article}

\usepackage{acl}

\usepackage{times}
\usepackage{latexsym}

\usepackage[T1]{fontenc}

\usepackage[utf8]{inputenc}

\usepackage{microtype}

\usepackage{colortbl}
\usepackage{booktabs}
\usepackage{graphicx}
\usepackage{paralist}
\usepackage{color}
\usepackage{array}
\usepackage{placeins}

\title{Investigating Cross-Domain Behaviors of BERT in Review Understanding}

\author{Albert Lu \\
  University of Notre Dame \\
  Notre Dame, IN 46556, USA \\
  \texttt{albertlu0310@gmail.com} \\\And
  Meng Jiang \\
  University of Notre Dame \\
  Notre Dame, IN 46556, USA \\
  \texttt{mjiang2@nd.edu} \\}

\begin{document}
\maketitle
\begin{abstract}
Review score prediction requires review text understanding, a critical real-world application of natural language processing. Due to dissimilar text domains in product reviews, a common practice is fine-tuning BERT models upon reviews of differing domains. However, there has not yet been an empirical study of cross-domain behaviors of BERT models in the various tasks of product review understanding. In this project, we investigate text classification BERT models fine-tuned on single-domain and multi-domain Amazon review data. In our findings, though single-domain models achieved marginally improved performance on their corresponding domain compared to multi-domain models, multi-domain models outperformed single-domain models when evaluated on multi-domain data, single-domain data the single-domain model was not fine-tuned on, and on average when considering all tests. Though slight increases in accuracy can be achieved through single-domain model fine-tuning, computational resources and costs can be reduced by utilizing multi-domain models that perform well across domains. 

\end{abstract}

\section{Introduction}
\begin{figure*}[t]
    \centering
    \includegraphics[width=0.89\textwidth]{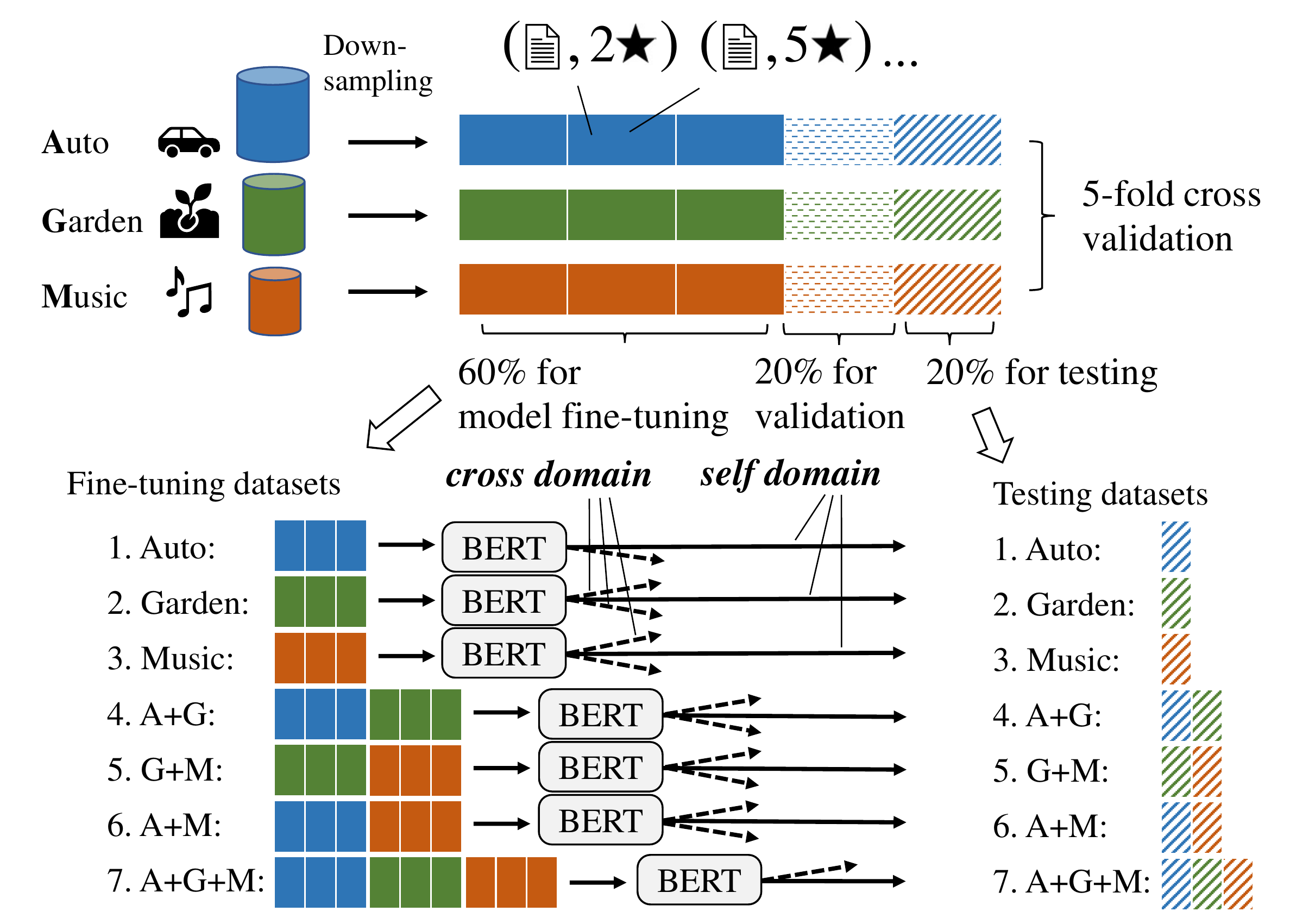}
    \caption{Seven datasets with varying combinations of domains were made, all of the same size. BERT models were trained upon all datasets and cross-domain tested. We investigate behaviors where the fine-tuning and testing domains are not fully aligned.}
    \label{fig:method}
\end{figure*}

Online services such as Amazon and Target have review systems where users can provide comments and ratings for products, offering valuable information to both buyers and companies. Analyzing the large volume of review data manually is impractical, so companies use data science and machine learning to extract knowledge efficiently ~\cite{alzubi2019analysis}. An important task in review understanding is review score prediction, akin to sentiment analysis, where the text of a review is used to determine the corresponding star rating. Review score prediction aids in distinguishing between the sentiment conveyed by the star rating and the sentiment expressed in the text, addressing potential biases and ensuring accurate ratings ~\cite{haque2018sentiment, elmurngi2018unfair}. 

Bidirectional encoder representations from transformers (BERT) models, a model set for natural language understanding, can be fine-tuned to predict star ratings based on review text~\cite{kenton2019bert}. BERT utilizes a transformer architecture, processing an entire text sequence at once, capturing contextual relationships between words. BERT consists of stacked transformer encoder layers with self-attention mechanisms and is pre-trained on two tasks: masked language modeling and next sentence prediction. After general pre-training, BERT is fine-tuned for specific downstream tasks by adding task-specific layers and adjusting neural network weights based on labeled data~\cite{kenton2019bert}, where rating score prediction is one of the popular downstream tasks.

The existence of multiple domains poses a significant challenge for review score prediction. Platforms like Amazon have diverse product categories, requiring models to analyze reviews for each domain separately. Though cross-domain models have been created, it remains unclear whether a single model can perform optimally across all domains ~\cite{liu2021embeddings, almoslmi2017crossdomain}. To address this, BERT can be fine-tuned using training sets specific to a domain or a mixture of training sets from various domains~\cite{sushil2021yet}. Fine-tuning with domain-specific data allows models to focus on predicting the relevant data they encounter, avoiding the counterproductive task of predicting irrelevant domains. However, training individual domain-specific models increases resource use compared to training a single model for multiple domains, assuming the multi-domain model performs well in different domains. To understand the trade-offs between a specific versus general approach for multiple domains, an empirical study analyzing BERT's cross-domain behaviors in review understanding is necessary.

We use data from three Amazon review domains to form seven datasets, fine-tuning from them seven BERT models. Each BERT model was evaluated upon every dataset in a round-robin format. Figure~\ref{fig:method} and the Methods section describe our study design with details about data processing and model fine-tuning. 

Below briefly are our observations. Predictions made before experimentation included that, though it would be possible to fine-tune BERT models to specialize on a certain domain, the benefit from single-domain fine-tuning would be minor. Also predicted was that multi-domain models would achieve a higher performance when evaluated on single-domain datasets compared to single-domain models not trained upon that specific domain. Observations lined up with predictions. Though BERT models were able to achieve domain specificity, the gains in performance towards single domains 
were slight. Moreover, the specificity gained appears to have decreased the single-domain models' ability to predict on other domains. Multi-domain models outperformed single-domain models on all categories besides the specific domain the single-domain models had been fine-tuned on, as well as achieving a higher average accuracy across all datasets. In addition, multi-domain models performed better on the multi-domain datasets the models were trained with compared to the single-domain models and their corresponding single-domain datasets. Towards applications in review analysis, review score prediction, and sentiment analysis, it appears that, though marginal performance improvements can be achieved through single-domain, specific fine-tuning, computational resources can be saved through training cross-domain models that achieve a high level of performance for all domains.

\section{Method of Our Study}
\begin{table*}[t]
    \centering
    \begin{tabular}{l|ccccccc|r}
    \toprule
    Model & \multicolumn{7}{c|}{Model was evaluated on test sets in: (\textbf{RMSE: smaller is better})} & \\
    (was trained in) & Auto & Garden & Music & A+G & G+M & A+M & A+G+M & Average \\
    \midrule
    Auto & \underline{0.640} & 0.742 & 0.781 & \cellcolor{blue!15}0.678 & \cellcolor{red!15}0.721 & \cellcolor{red!15}0.632 & \cellcolor{red!15}0.711 & {0.701} \\
    Garden & \textbf{0.624} & \underline{0.721} & 0.854 & \cellcolor{red!15}0.707 & \cellcolor{red!15}0.787 & \cellcolor{red!15}0.648 & \cellcolor{red!15}0.724 & {0.724} \\
    Music & 0.860 & 0.781 & \underline{0.728} & \cellcolor{red!15}0.860 & \cellcolor{red!15}0.714 & \cellcolor{red!15}0.806 & \cellcolor{red!15}0.783 & {0.790} \\
    A+G & \cellcolor{red!15}0.721 & \cellcolor{blue!15}0.707 & \cellcolor{red!15}0.794 & \underline{0.686} & 0.643 & 0.632 & 0.663 & {0.691} \\
    G+M & \cellcolor{red!15}0.714 & \cellcolor{blue!15}\textbf{0.700} & \cellcolor{red!15}0.794 & 0.686 & \underline{\textbf{0.640}} & 0.632 & 0.661 & {0.689} \\
    A+M & \cellcolor{red!15}0.748 & \cellcolor{red!15}0.806 & \cellcolor{blue!15}\textbf{0.707} & 0.735 & 0.693 & \underline{\textbf{0.616}} & 0.731 & {0.719} \\
    A+G+M & \cellcolor{red!15}0.678 & \cellcolor{blue!15}0.714 & \cellcolor{blue!15}0.714 & \textbf{0.663} & 0.641 & 0.624 & \underline{\textbf{0.660}} & \textbf{0.670} \\
    \midrule
    Average & {0.712} & {0.739} & {0.767} & {0.716} & {0.691} & {0.656} & {0.704} & Best avg: 0.659 \\
    \bottomrule
    \end{tabular}
    \caption{Results from a comprehensive investigation on cross-domain review score prediction using BERT models.}
    \label{tab:results}
\end{table*}

\begin{table*}[t]
    \centering
    \begin{tabular}{l||c|c|cc|cc|c}
    \toprule
    Model & Self & All & \multicolumn{2}{c|}{Specific} & \multicolumn{2}{c|}{Semi-Specific} & General \\
    specificity & & & Familiar & Unfamiliar & Familiar & Unfamiliar & \\
    \midrule
    Specific & 0.696 & 0.739 & \textbf{0.696} & \textbf{0.774} & 0.721 & 0.789 & 0.739 \\
    General & \textbf{0.649} & \textbf{0.698} & 0.721 & 0.754 & \textbf{0.667} & NA & \textbf{0.697} \\
    \bottomrule
    \end{tabular}
    \caption{Table of average RMSEs of models towards various groups of datasets.}
    \label{tab:specific}
\end{table*}

Three Amazon review datasets with mutually dissimilar domains were chosen to fine-tune BERT with~\cite{mcauley2015image,he2016ups}:
automobile part review data (Auto); musical instrument review data (Music); and patio, lawn, and garden review data (Garden). Datasets consisted of examples with an average review text length of 582 characters and a one-to-five rating.

Our pipeline process from data sampling to model evaluation is shown in Figure~\ref{fig:method}.
Data for model training was processed through downsampling to reduce domain and class imbalance. Since the original Auto domain data is significantly larger than the other datasets, 2500 reviews were sampled from each domain for model fine-tuning, remedying domain imbalance. To reduce class imbalance, for each domain, the 2500 training reviews consisted of 200, 250, 300, 600, and 1150 one- to five-star reviews, respectively. 800 reviews were randomly sampled from each domain for both validation and test splits. 

Processing resulted in a final training corpus of 7500 reviews, divided evenly among the three domains. Four more multi-domain datasets were created through combining examples from various domains. Three two-domain datasets with 2500 train examples each were formed through taking 1250 examples from each combination of two single-domain datasets. One three-domain dataset of 2500 train examples was formed from an even mixture of examples from all three domains. 

The pre-trained \emph{BERT-base-cased} model served as the fine-tuning backbone. Batches of fine-tuned models configured with varying hyperparameters were trained for each dataset to discover optimal model performance towards a specific dataset. Four learning rates (5e-5, 4e-5, 3e-5, and 2e-5) and four batch sizes (32, 64, 128, and 256) were used along with a 20 epoch training duration and a learning rate warmup ratio of 0.1. High learning rates and batch sizes were found to increase model performance, and after selecting the best model from each batch, each of the seven models was evaluated against the test split of every dataset. 

The root mean squared error (RMSE) metric was used to determine the optimal model in each dataset batch and evaluate model cross-domain testing. As RMSE is proportional to the difference between true and predicted values, a smaller RMSE indicates higher performance. 

Models were trained and tested on a GPU server of four NVIDIA GeForce RTX 2080 Ti cards, managed by the Center for Research Computing at the University of Notre Dame.





\section{Results and Analysis}
\subsection{Model-level results and analysis}

The results of testing each model against each dataset are shown in Table~\ref{tab:results}. Every row of the table represents one model being tested on all seven datasets; every column represents one dataset having all seven models evaluated upon it. Performance of the models was measured using Root Mean Squared Error (RMSE). A smaller RMSE value indicates higher model accuracy. Bold table values represent the highest performance achieved in each column. The ``Average'' across each column and each row indicates the mean score for that column or row. Considering the horizontal row of averages, higher RMSE values indicate a more difficult dataset. Considering the vertical average column, lower RMSE values mean a better-performing model. The ``Best avg'' in the bottom right corner is found by taking the best (lowest) RMSE for each test set and calculating the mean of those values. 0.659, the best average, is quite good considering the rating score ranges from one to five. Within-domain performance (Auto model with Auto data) is indicated on the table through underlining. The bottom left cells contain the RMSEs of multi-domain models tested on single-domain data, while in the top right the RMSEs of single-domain models on multi-domain data are shown. Red cells mean a higher RMSE than the underlined baseline value of the column; blue cells indicate the inverse. Our observations are as follows.

\textbf{(1) Multi-domain models perform better than single-domain ones averaged across single- and multi-domain datasets.} The three-domain A+M+G model achieved the lowest RMSE, 0.670, averaged across all seven datasets. This matches intuition - though single-domain models were expected to and did perform better on certain specific domains, overall the more cross-domain a model was the lower the error across all domains. This appears to be due to the multi-domain models having received exposure to data from varied domains, and so having a ``background'' to predict decently well on music, auto, and garden data, whereas single-domain models, though having high performance on the single domain it was fine-tuned upon, lack knowledge of domains outside its ``scope,'' and so performed poorly on those. It seems that the decreases in performance for single-domain models on unfamiliar domains outweighed the high accuracy on the sole familiar domain, causing single-domain models to be inferior to multi-domain models on average. 

\textbf{(2) Some domains are more difficult for models to predict well upon, perhaps due to differences in text content and technicality.} On average, models when evaluated on the Music dataset resulted in the worst performance, with a RMSE of 0.767 averaged across all seven models. Perhaps this was due to the high musical knowledge of those who purchase musical instruments causing reviews to contain more difficult vocabulary. One would think automobile and garden products (on Amazon) would not be often purchased by mechanics or professional gardeners but by amateurs whose product reviews would not utilize technical language. Musical instruments, however, are in general more expensive and niche than a simple spark plug or lawn gnome and so reviews would tend towards professionals who, instead of merely commenting with ``good violin'' or ``bad violin'' would use terms that may be clear to a human musician but would be difficult for a BERT model to interpret as positive or negative.  The easiest domain for models seems to be the automobile part data, with an average RMSE of 0.712. With regards to the Auto data, an outlier presents itself: the garden model, towards the auto domain, had a higher performance than even the automobile model itself. Given the uniqueness of this outlier (there is no other example in the table of a model completely unfamiliar with a domain outperforming a model trained upon that domain), this unexpected high-performing result can be treated as a curiosity caused by data variation between domains, interesting but ultimately unimportant to overall analysis.

\textbf{(3) Data combining domains allows a wider range of models to predict well upon it.} Looking at the latter four columns of Table 1, it seems that on average models achieved a lower RMSE when evaluated with multi-domain data (two- and three-domain data). This should not be taken as a sign that difficult domains when combined result in an easier dataset. The relative higher performance of models towards multi-domain data should be taken as more models having familiarity towards a section of the multi-domain data and thus, overall, performing better. 

\textbf{(4) Multi-domain models outperformed single-domain models in all multi-domain data, and can outperform single-domain models in their corresponding single-domain dataset, perhaps caused by the ability of multi-domain models to adapt to variations in data.} These interesting observations can be found by considering the blue-and-red shaded top right and bottom left regions of Table 1. First, the top right, single-domain model with multi-domain dataset, section had only one blue cell - only one model-dataset pair that achieved a better RMSE than the baseline, underlined value in that column. It seems that, when considering multi-domain data, no single-domain model can compete with the multi-domain model. However, through analyzing the bottom left table quadrant, the inverse turns out to be untrue. Multiple cells are blue; considering the garden and music datasets, more than half of the multi-domain model with single-domain dataset tests resulted in a higher performance than the baseline of the column. It seems that, in the auto column, due to variations in data domains discussed above, the single-domain auto model was able to achieve a high enough performance that no multi-domain model was able to exceed its accuracy. In the garden and music domains, in both cases the three-domain model outperformed the baseline single-domain model, and in the garden domain, both of the two-domain models that had garden data as part of their fine-tuning outperformed the baseline. For the music dataset, the auto plus music model was the best of its column. This is unintuitive, since one would expect each single-domain model to have the highest performance towards its corresponding domain. This may have been caused by subtle effects produced by the difference in training versus testing data. Undersampled training data is not representative of a real-world application; however, fine-tuning a model with class-biased, unprocessed amazon reviews creates a model that only outputs fives and fours. The single-domain models may have been exposed to such homogeneous data that the model was overfitted to the processed, undersampled dataset, and thus could not predict well on the test data. Multi-domain models may have avoided this issue by heterogeneous, multi-domain data (still undersampled) allowing the model to be exposed to ``variations'' in data. In the training data, the variation is solely a difference in domain; in evaluation, the testing data both has a difference in domain between examples and a difference in class bias compared to the training data. Multi-domain models are more accustomed to variation than single-domain models due to the nature of multi-domain data; it is not a stretch to deduce that multi-domain models could better accommodate the differences between training and testing data. 

\subsection{Specific models vs general models}

Table~\ref{tab:specific} summarizes average RMSEs among specific versus general models for differing categories of dataset. Specific is defined as single-domain; general models are the four two- and three-domain models. The ``Self'' column corresponds to the underlined values in Table~\ref{tab:results} - the RMSE of each model tested against its corresponding domain. The ``All'' column is the average of the specific or general models' RMSEs on all datasets. In Table~\ref{tab:specific}, ``Specific'', ``Semi-Specific'', and ``General'' datasets are defined as single-domain, two-domain, and three-domain datasets respectively. The ``Specific'' and ``Semi-Specific'' columns are divided into two sub-column - ``Familiar'' and ``Unfamiliar.'' Familiar is defined as a model having had any training on any constituent domain of a dataset - for example, the Auto dataset would be familiar to both the ``Specific'' Auto model and the ``General'' A+G model, but would be unfamiliar to the M+G model. (The ``NA'' under ``Semi-Specific, Unfamiliar'' is due to all three two-domain datasets being familiar to all two- and three-domain models, as all pairs of two-domain dataset and two- or three-domain model share at least one domain.) The two values in the ``General'' column are the average of the RMSEs of all three single-domain models tested on A+M+G and the average of the RMSEs of all four multi-domain models tested on A+M+G, respectively. 

Most of the conclusions found by analyzing the results displayed in Table~\ref{tab:results} can also be found in Table~\ref{tab:specific}. Though general models had a lower RMSE than specific ones across all datasets, single-domain models, on average, still performed better on their corresponding domain compared to multi-domain models on both familiar and unfamiliar specifc data; 0.696 compared to 0.721 and 0.754, respectively. However, other interesting conclusions can be found by further analyzing Table~\ref{tab:specific}. 

\begin{table*}[t]
    \centering
    \begin{tabular}{p{0.1\linewidth}|p{0.5\linewidth}|p{0.05\linewidth}| p{0.1\linewidth} | p{0.1\linewidth}}
    \toprule
    Domain & Review Text & True & Model & Predicted \\ 
    \midrule
    Music & I'm another reviewer who has, upon personal inspection, discovered that the rear leg of the stand is too short to provide staedy support for any beloved guitar. I returned it. I've First Act brand stands similar in appearance but more intelligently constructed. & 1 & Music \newline Auto \newline A+M & 3 \newline 4 \newline 2 \\
    \midrule
    Garden & This sprayer is very well built and solid. It is easy to use and with the long hose and 18 inch wand it sprays a very good distance and the spray is of course adjustable to different spray patterns. I am happy with hit. & 5 & Music \newline Auto \newline Garden & 5 \newline 5 \newline 5 \\
    \midrule
    Auto & Takes the dust off my car without leaving any streaking that some report with other brands. Just don't press real hard. & 5 & Garden \newline Auto \newline A+G & 3 \newline 5 \newline 5 \\
    \bottomrule
    \end{tabular}
    \caption{Examples of reviews from various domains with review text, true rating, and predicted rating by multiple models, illustrating several conclusions detailed in the Case Study section.}
    \label{tab:examples}
\end{table*}

Despite Table~\ref{tab:results} indicating that in some cases multi-domain, general models could outperform single-domain models, \textbf{(5) in general multi-domain models tended to lose accuracy when evaluated on data with a single domain.} Compared to the baseline self-test average RMSE of 0.649, the general models had a 0.072 and 0.105 increase in RMSE when evaluated on familiar and unfamiliar single-domain data, respectively. A decrease in the performance of multi-domain models on single-domain data is also apparent when comparing the ``Specific'' column with the ``Semi-Specific'' and ``General'' columns of Table~\ref{tab:specific}, specifically the values in the ``General'' row. The average RMSEs of 0.667 and 0.697 of general models towards two- and three-domain data, respectively, are superior to both 0.721 and 0.754, the RMSEs of familiar and unfamiliar specific data, respectively. The loss of performance on unfamiliar data is expected; however, lower accuracy on familiar single domains is more interesting. An example of lower accuracy of a multi-domain model on familiar single-domain data is the A+M model achieving a worse RMSE on the Auto dataset and the Music dataset compared to the A+M dataset. This lowered performance on familiar domains in isolation may be caused by multi-domain models, in general, being unable to fully predict well on single-domain data that may contain technical vocabulary that, though familiar, has not been completely fine-tuned to. Single-domain test datasets are also more prone to outliers, at the data sample size used - for example, a certain subsection of Auto data, selected for testing, may contain a greater-than-average level of reviews utilizing technical vocabulary; multi-domain datasets reduce the chances of a test set with complicated vocabulary the multi-domain model cannot predict well upon. Regardless of the cause, however, multi-domain models achieved a relatively lower performance on single-domain data, compared to the self-test and testing on both two- and three-domain data. 

A conclusion made when analyzing Table~\ref{tab:results} is further supported by the averages in Table~\ref{tab:specific}. \textbf{(6) Specific models, when tested on general data and specific data not of the model's domain, exhibited an expected decrease in performance.} Compared to the baseline self-test average of 0.696 among the three single-domain models, the average RMSEs of the single-domain models tested on familiar two-domain, unfamiliar two-domain, and three-domain datasets were 0.025, 0.093, and 0.043 higher, respectively. A decrease in performance towards unfamiliar specific data was also observed, the increase in RMSE being 0.075 there, similar to the 0.093 value mentioned above for the increase in RMSE with specific models tested on unfamiliar semi-specific data. Note that the relatively lower performance decreases in familiar two-domain and three-domain datasets is very probably not caused by Auto or Garden models predicting better on the Music-domain half of the A+M dataset compared to the Music data alone. The higher accuracy of the Garden model towards A+G data versus A+M data is because fifty percent of the former dataset is Garden-domain, while the Garden domain is not present in the latter dataset. Single-domain model only perform worse on semi-specific and general datasets because of the unfamiliar specific data content of the multi-domain datasets.

\subsection{Case study}

Some examples that highlight the conclusions presented above are shown in Table~\ref{tab:examples}. Echoing points 2, 4, and 7, example 1 shows a review which all models failed to predict well upon (Music, Auto, and A+M are shown to highlight various conclusions). This review showcases the varied aspects of Music domain reviews that may have created difficulty for model prediction - lack of clear adjectives and a sentiment that requires knowledge of what ``guitars", ``rear legs", and ``stands" are. The model predictions also support that single-domain models lose accuracy when tested on a different domain (the Auto model predicted a 4), and that, in some cases, a two-domain model can outperform a single-domain one (A+M performance). 

Example 2 provides an example of an ``easy" review to predict upon. The presence of ``very good" and ``happy with hit [sic]" allows all models to predict well upon the review. It also highlights a point that aids cross-domain models. As one can see in this example, ``simple" adjectives such as ``good" and ``well" are present in great amount, a trait common to most reviews. Models undergoing cross-domain testing can utilize these keywords (Auto and Music model, which both predicted the true value) to achieve accuracy on various domains. 

Example 3 provides another review that contains domain-specific text with few ``simple" adjectives. The Garden model may have latched onto the ``don't" and ``without", misinterpreting those negative-connotation words as indicative of a low star rating. Example 3 again shows that single-domain models predict well on the own dataset, and that multi-domain models with familiarity towards a certain domain can predict just as well. 

\paragraph{Summary:}
To summarize the conclusions presented through analysis of Table~\ref{tab:results} and Table~\ref{tab:specific}:
\begin{compactitem}
    \item On average, multi-domain models perform better than single-domain ones.
    \item Domains vary in difficulty of prediction. 
    \item Multi-domain data allows for both single- and multi-domain models to predict well.
    \item Multi-domain models outperform single-domain models in multi-domain data and can occasionally outperform single-domain models in the single-domain model's corresponding domain. 
    \item Multi-domain models lose accuracy when evaluated upon all single-domain data, including familiar data. 
    \item Single-domain models lose accuracy when evaluated upon unfamiliar single-domain data. 
    \item On average, single-domain models achieve higher accuracy towards their domain compared to multi-domain models.
\end{compactitem}

\section{Related Work}
\subsection{Domain adaptation and BERT}

Work has been done on cross-domain sentiment analysis with BERT~\cite{myagmar2019sentiment}. In the real world, the domain-dependent nature of sentiment analysis often creates issues when the amount of data that can be gathered for a specific domain is insufficient to properly train a high-performing, single-domain model. This problem is often solved through domain adaptation, in which a model is trained upon data from a relevant, secondary domain and deployed for the primary, data-scarce domain. Existing work has shown that fine-tuned BERT models outperform state-of-the-art domain adaptation methods~\cite{lin2020adapt}. The multi-domain models in our work show a similar capacity to adapt to varying domains; however, single-domain models were less capable of predicting well on unfamiliar domains \cite{wang2020calendar,wang2021modeling}. 

Another domain-reliant task is automatic speech recognition (ASR), in which it is common practice to pretrain generic ASR models and then fine-tune the models to domain-specific data ~\cite{bell2020asr}, an approach similar to BERT pretraining and fine-tuning. Other techniques for ASR domain adaptation include data augmentation, multi-task learning, teacher-student training, semi-supervised training, and using synthetic data created by a Text-to-Speech system ~\cite{bell2016augmentation, sun2017multitask, meng2019teacherstudent, synnaeve2020semisupervised, joshi2022asr}. If domain adaptability and the high performance achieved by single-domain models in our work are required, instead of multi-domain models the techniques used in ASR could be employed. 

\subsection{Knowledge distillation}
Work by Howell et al. on domain-specific knowledge distillation for conversational commerce is closely related to this work~\cite{howell2022distillation}. Knowledge distillation addresses the issue of large and computationally expensive models being too costly to be deployed through a "teacher-student" training approach, in which a trained "teacher" model facilitates the training of a smaller "student" model ~\cite{sanh2019distilbert}. Howell et al. show knowledge distillation can both reduce model size and adapt models to a specific domain. In this work, we show that BERT base models are capable of adapting to a domain through fine-tuning on domain-specific, instead of knowledge distillation. However, we also show that, compared to single-domain models, multi-domain models, though incurring a slight accuracy penalty, still retain high performance towards a broad range of data. We also propose multi-domain models as an alternative solution to cost of computation. Multi-domain models and knowledge distillation could be used in tandem to further reduce computational resources, using knowledge distillation to train a "student" model with a broad spectrum of data domains, creating a model that is both small in size and applicable to multiple domains.

\section{Conclusion and Future Work}
We investigated BERT models fine-tuned on single-domain and multi-domain data for the task of Amazon review score prediction. Though single-domain models achieved marginally improved performance on their corresponding domain compared to multi-domain models, multi-domain models outperformed single-domain models when evaluated on multi-domain data, single-domain data, and on average. Though slight increases in accuracy can be achieved through single-domain model fine-tuning, computational demands and costs can be reduced by utilizing multi-domain models that perform well across many domains. 

\subsection{Discussion and impact}

The primary goal of this work was to determine the tradeoffs of specific versus general model domain scopes in BERT review score prediction. The difference between model RMSEs was comparatively low relative to the raw RMSE values, around 5\% for most variations. This may have been caused by the task of classifying Amazon reviews not requiring much fixation on topic-related words. Though soap and furniture reviews contain differing nouns, adjectives for negative and positive soap and furniture reviews would be mostly alike. The BERT models may have recognized the importance of sentiment-laden adjectives such as ``bad'' or ``good'' that can apply to any topic product when determining the review's star rating, placing weight on more general keywords rather than topic-specific nouns and descriptions. However, review score prediction still requires a certain level of domain understanding, supported both by our results and by phrases such as "takes up all my time," which is negative when referring to furniture setup but positive when applied to books. 

The findings in this work support that specific models are still superior to general models for specific applications. Tasks that come to mind that specific models would be best suited for include, say, diagnosis of a certain cancer type through processing a patient's symptom log or as a search aide for an online archive containing papers of a narrow breadth. However, these specific models will necessarily be limited in scope.

Though general models in this predictive task were still capable of achieving relatively high performance on all domains, they lost accuracy on specific data, even data with domains that the general model had been exposed to before. Search engines and chatbots come to mind as tasks that require a general solution. However, general models often fail to correctly interpret a specific instruction. For example, a text summarization service may not be able to interpret a technical text accurately.

A possible solution for the above would be to utilize a general NLP model in order to determine what domain a certain piece of text falls into – a relatively simple task that is suited to a broad, general model. The text could then be sent to a specific model customized to address the particular domain the general model determined the text fell under. This system, though resource-intensive, would both allow the general model to perform a task that does not require the precision of a specific model and prevent the specific models from encountering any text outside its domain breadth.

\subsection{Further experimentation}

Further experimentation along these lines could include expanding the number of datasets used in order to minimize the effects outliers or poorly fine-tuned models have on results. Such experimentation would also be able to determine how the ``interrelatedness'' of topics affects BERT and NLP – one would expect soap and detergent models to be more similar than soap and furniture models. Single- versus multi-domain testing could also be done upon tasks requiring NLP models to place weight on topic-specific vocabulary (such as text summarization or keyword extraction), creating a true ``specific'' model, avoiding the problem described in the Discussion section of specific models fixating on general adjectives. Such investigation would be able to determine the tradeoffs between general and specific scopes in the broader, NLP field, and whether the trends detailed in this work regarding review score prediction with BERT (a narrow subset of NLP) do or do not extend to the rest of the e-commerce and NLP field.

\section*{Acknowledgement}

This work was supported by National Science Foundation (NSF) IIS-2119531, IIS-2137396, IIS-2142827, IIS-2234058, CCF-1901059, and Office of Naval Research (ONR) N00014-22-1-2507.

\newpage

\bibliography{main}
\bibliographystyle{acl_natbib}




\end{document}